\documentclass[acmsmall]{acmart}

\usepackage{CJKutf8}
\usepackage{bm}
\usepackage{multirow}
\usepackage{graphicx}
\usepackage{epstopdf}
\usepackage[subfigure]{tocloft}
\usepackage{subfigure}
\usepackage{array}

\AtBeginDocument{%
  \providecommand\BibTeX{{%
    \normalfont B\kern-0.5em{\scshape i\kern-0.25em b}\kern-0.8em\TeX}}}

\begin{document}
\setcopyright{acmcopyright}
\acmJournal{TALLIP}
\acmYear{2022} \acmVolume{1} \acmNumber{1} \acmArticle{1} \acmMonth{1} \acmPrice{15.00}\acmDOI{10.1145/3426882}
\title{Few-shot Incremental Event Detection}


\author{Hao Wang}
\email{wanghaomails@gmail.com}
\author{Hanwen Shi}
\email{hanwenwork@163.com}
\author{JianYong Duan}
\email{duanjy@ncut.edu.cn}
\authornote{Corresponding authors}
\affiliation{%
  \institution{School of Information Science and Technology, North China University of Technology, and CNONIX National Standard Application and Promotion Lab}
  \streetaddress{ShiJingShan district JinYuanZhuang No.5, BeiJing, 100144, China}
 }






\begin{abstract}
Event detection tasks can enable the quick detection of events from texts and provide powerful support for downstream natural language processing tasks.
Most such methods can only detect a fixed set of predefined event classes. 
To extend them to detect a new class without losing the ability to detect old classes requires costly retraining of the model from scratch. 
Incremental learning can effectively solve this problem, but it requires abundant data of new classes. 
In practice, however, the lack of high-quality labeled data of new event classes makes it difficult to obtain enough data for model training. 
To address the above mentioned issues, we define a new task, few-shot incremental event detection, which focuses on learning to detect a new event class with limited data, while retaining the ability to detect old classes to the extent possible.
We created a benchmark dataset IFSED for the few-shot incremental event detection task based on FewEvent and propose two benchmarks, IFSED-K and IFSED-KP.
Experimental results show that our approach has a higher F1-score than baseline methods and is more stable.
\end{abstract}




\maketitle

\begin{CJK*}{UTF8}{gbsn}
\section{Introduction}
\label{sec:introduction}

\begin{figure*}[!t]
    \centering
    \includegraphics[width=0.89\linewidth]{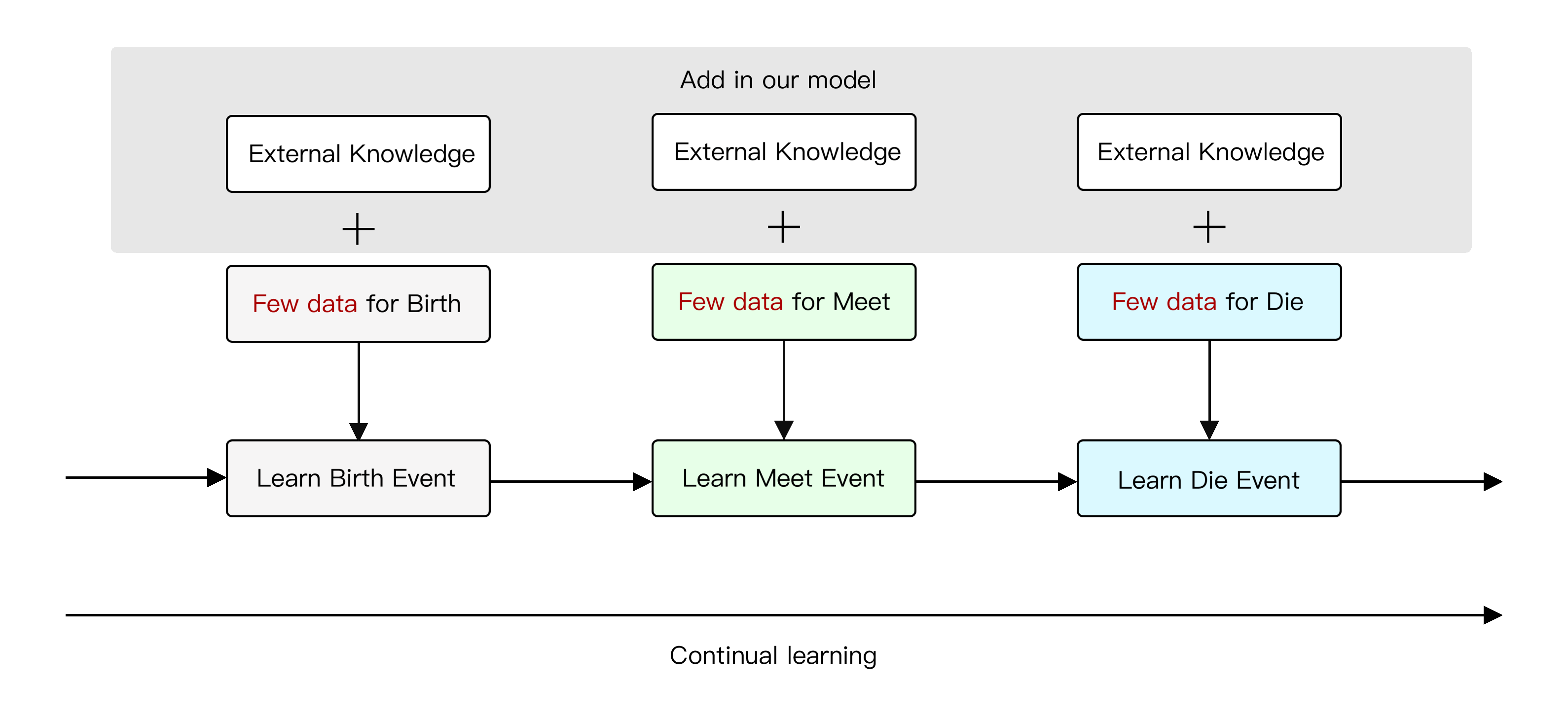}
    \caption{Few-shot Incremental Event Detection: a model must learn new samples while retaining the ability to recognize previous samples. For example, the model should learn the "meet event" class using its external knowledge and a few samples of the class. After learning the meet event class, the model should be able to detect both meet events, as well as birth events, which it learned previously.
}
    \label{fig:IFSED}
\end{figure*}

Texts are increasingly part of daily life, whose sentences contain abundant information \cite{2020-Small-survey}, which we wish to quickly extract.
Event detection aims to detect trigger words and classify them into specific event types \cite{2021-AKE-FED}. 
This plays an important role in the monitoring of public opinion, emergency alerts, and intelligence gathering \cite{2019-meta-Extractor-survey}.
To improve the accuracy of event detection can provide powerful support for downstream tasks.

Most event detection methods can only detect predefined types of information in a world where new event types emerge constantly.
Computational costs preclude the retraining of an event detection model each time a new event class appears.
An event detection system should be able to incrementally learn new event classes.
Introducing incremental learning into event detection is an effective way to solve this problem. 
However, incremental learning often requires abundant data of a new class, and to collect enough labeled data is usually impossible \cite{2021-AKE-FED}.

Few-shot learning uses only a few (e.g., 5 or 10) labeled samples for learning \cite{2020-MetaFS-survey}, greatly reducing learning costs, and can help   solve this problem. 
We define the few-shot incremental event detection task, as shown in Fig. \ref{fig:IFSED}, which is more realistic and complex than few-shot and incremental event detection tasks. 
This task focuses on learning new classes without retraining from scratch and losing the ability to recognize old classes.
We constructed a benchmark dataset for few-shot incremental event detection. We cleaned up the FewEvent dataset and divided the data into base,  incremental, and out-of-distribution (OOD) data.
We split the incremental data into five rounds to evaluate the incremental learning ability of the proposed models. 
In each round, data are divided in 5-way-5-shot (five classes in each round of training, with five samples per class) and similarly defined 10-way-10-shot to explore the effects of different numbers of classes and sample sizes.

We make the following contributions:

1) We propose the few-shot incremental event detection task, which focuses on learning the ability to detect new event classes from limited labeled data of new event classes;

2) We reconstruct a benchmark dataset IFSED for the few-shot incremental event detection task based on the FewEvent dataset \cite{2002-FrameNet-BFP}. 

3) We propose the IFSED-K and IFSED-KP models, which outperform other models according to experimental results.

\section{Models}
\label{sec:models}

\subsection{Problem Formulation}
In this section, we formalize the few-shot incremental event detection task. 

\noindent\textbf{Training data}

In the setup of few-shot incremental event detection, the system provides m rounds of new classes $\lbrace C_{1}, ... , C_{m}\rbrace$ in order.
Each round of $C_{i}$ has n new classes with $C_{i}=\lbrace C_{i}^{1}, \dots , C_{i}^{n}\rbrace$.
Each new class has only k samples (k=5 or 10), and the value of k is fixed. 
To set additional base classes, the system can access a set of base classes $C_{b}=\lbrace C_{b}^{1}, \dots , C_{b}^{g}\rbrace$. 
All base classes $C_{b}$ have enough labeled samples for training. The number of all base classes is g.
To mimic a real-world situation, we create a multi-round setting, feeding the model with new classes in each round in order. 
We set k to 5 or 10, and k should be consistent for each experiment.

\noindent\textbf{Development data}

In few-shot incremental event detection, only k available samples are set for each new class.
We use no development dataset to help model training.
We choose hyperparameters such as batch size based on suggestions by Hugging-face and others, such as Cao \cite{2020-KCN-IED} and Shen \cite{2021-AKE-FED}. 
We modify hyperparameters for experiments by training data and select the best ones.
However, we will introduce the development set in subsequent studies.
We recommend that hyperparameters be selected based on experience or relevant experiments.

\noindent\textbf{Testing data}

The testing data consist of examples across all classes, i.e., the base, new, and OOD classes. 
The test data are $C_{1}^{'} \cup \cdots \cup C_{m}^{'} \cup C_{b}^{'} \cup C_{o}^{'}$, where $C_{b}^{'}$ is the base class, $C_{o}^{'}$ denotes the OOD classes, and $C_{b}^{'}$ and $C_{o}^{'}$ are options that can be used in the test data.
OOD classes refer to the relevant samples that the model does not learn.
OOD classes test the ability of the model to detect new classes that have not been learned.
These are essential, as incremental learning of open learning problems encounters unlearned new classes in practice.

\noindent\textbf{Requirements}

We focus on the performance in different rounds and classes, including the base class and new classes in different rounds.
We expect that this model can learn new knowledge by continuous learning on samples.
At the same time, new classes learned in previous rounds should not be forgotten.
An effective model can be applied in real life and should not have catastrophic forgetting problems.

\noindent\textbf{Comparison}

We use Table \ref{table:different} to show their differences between incremental event detection, few-shot event detection and few-shot incremental event detection.

\begin{table}[]
\begin{tabular}{|>{\centering\arraybackslash}m{3cm}|>{\centering\arraybackslash}m{3cm}|>{\centering\arraybackslash}m{3cm}|>{\centering\arraybackslash}m{3cm}|}
\hline
& Few-shot event detection   
& Incremental event detection   
& Few-shot incremental event detection   \\ \hline 
Are the training data few-shot samples?   & Yes   & No   & Yes  \\ \hline
Can models detect unlearned class?      
    & No.
    & Yes, by incremental learning.                                                  
    & Yes, by incremental learning.\\ \hline
Focus 
    & The inputs are few-shot data.   
    & Models can incrementally learn new classes.
    & Models can incrementally learn new classes, but only need few samples. \\ \hline
\end{tabular}
\caption{The difference between the three tasks.}
\label{table:different}
\end{table}

1) Few-shot event detection: 
Few-shot event detection is a challenging task that is typically framed as an N-way-K-shot problem, whereby a tiny labeled support set S is provided for model training. 
Specifically, S consists of N distinct event types, each of which has only K labeled samples, where K is often a small number (e.g., K=5 or K=10). 
The main of this task focus on few-shot data on input.
Notably, the model is unable to detect unlearned classes. 
While meta-learning has been proposed as a means to detect such classes, the accuracy of this approach remains insufficient for practical applications. 
The main focus of meta-learning in few-shot learning is on few-shot, generally by training the model on the base class and then generalizing it to the new class.

2) Incremental event detection: 
Incremental event detection requires a model to learn new classes incrementally while retaining the ability to detect old classes. 
To achieve this goal, models must focus on incrementally learning new classes, but this requires a sufficient number of high-quality samples for learning.

3) Few-shot incremental event detection: 
Few-shot incremental event detection presents an even more challenging problem, as it requires a model to learn new event classes with limited data while retaining the ability to detect old classes to the best extent possible. 
Models are able to perform incremental learning with only a small amount of data. 
In few-shot learning, the focus is on incremental learning, whereby the model must quickly learn from a few new samples without forgetting what it has already learned.

\subsection{Our models: IFSED-K and IFSED-KP}
To conduct the study, we constructed models based on two learning methods.
The few-shot incremental event detection model with external knowledge (IFSED-K) uses a general classifier to detect and learn event classes, and the few-shot incremental event detection model with external knowledge and prototype network (IFSED-KP) uses a meta-learning approach to detect and learn event classes by teaching the model how to learn. 
\begin{figure*}[!t]
    \centering
    \includegraphics[width=0.89\linewidth]{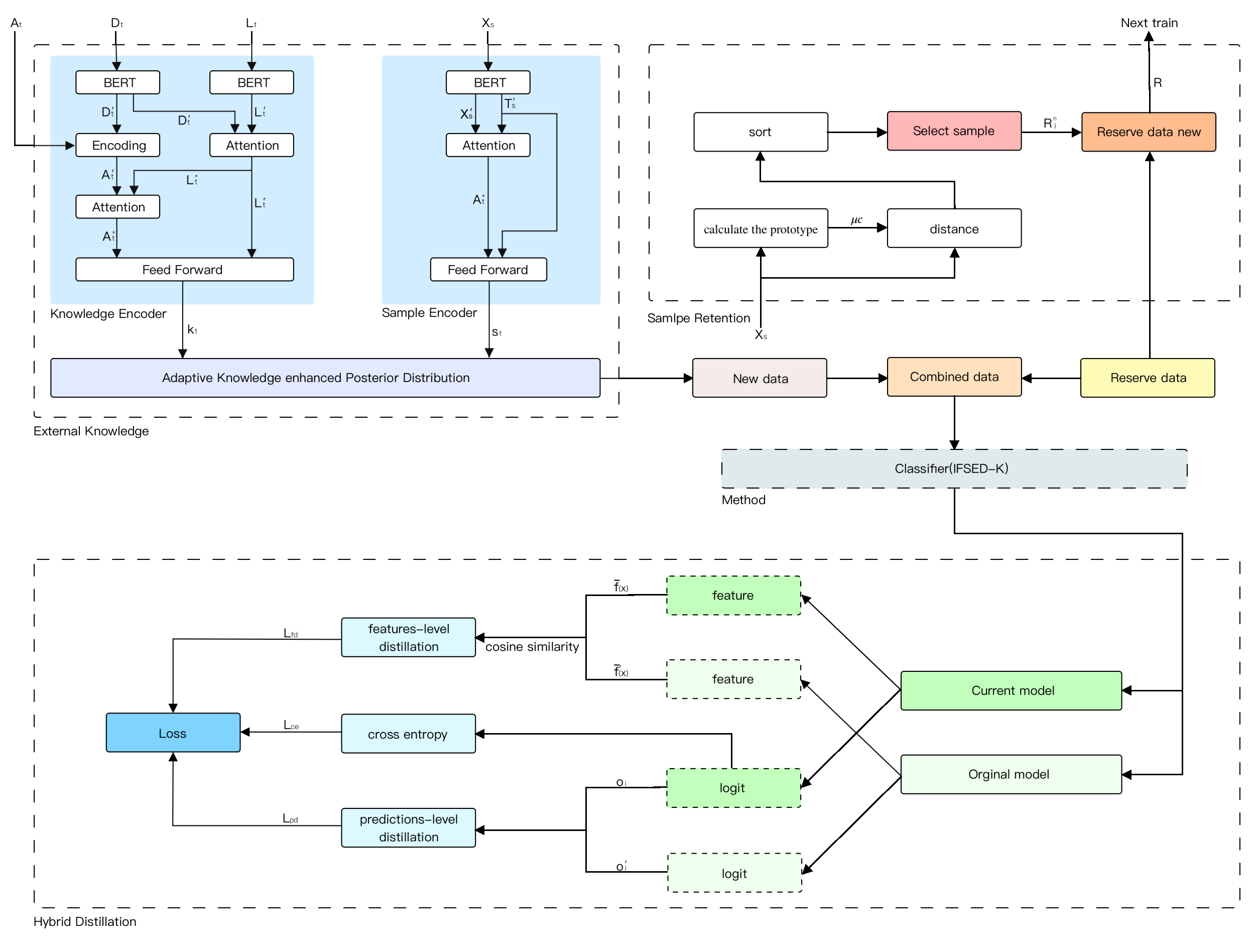}
    \caption{Framework of IFSED-K model: 1) external knowledge; 2) sample retention; 3) model construction and training; 4) hybrid distillation}
    \label{fig:IFSEDK}
\end{figure*}

As shown in Figs. \ref{fig:IFSEDK} and \ref{fig:IFSEDKP}, both methods have four steps, where steps 2 and 4 are the same, and steps 1 and 3 differ. 

1) External knowledge: When a new round appears, the model obtains the lexical units, arguments, and definitions from FrameNet and encodes them using the knowledge encoder and sample encoder.
We do not use an adaptive knowledge-enhanced Posterior in IFSED-KP;

2) Sample retention: After a new event class is trained, its most representative sample is selected, saved, and added to the next training session;

3) Model construction: The coding information obtained in the first step, the new and retained samples obtained in the second step are respectively trained with the classifier (IFSED-K) and meta-learning (IFSED-KP) methods;

4) Hybrid distillation: We calculate the loss function by hybrid distillation.

\subsubsection{\textbf{External Knowledge}}
\ 

\begin{figure*}[!t]
    \centering
    \includegraphics[width=0.89\linewidth]{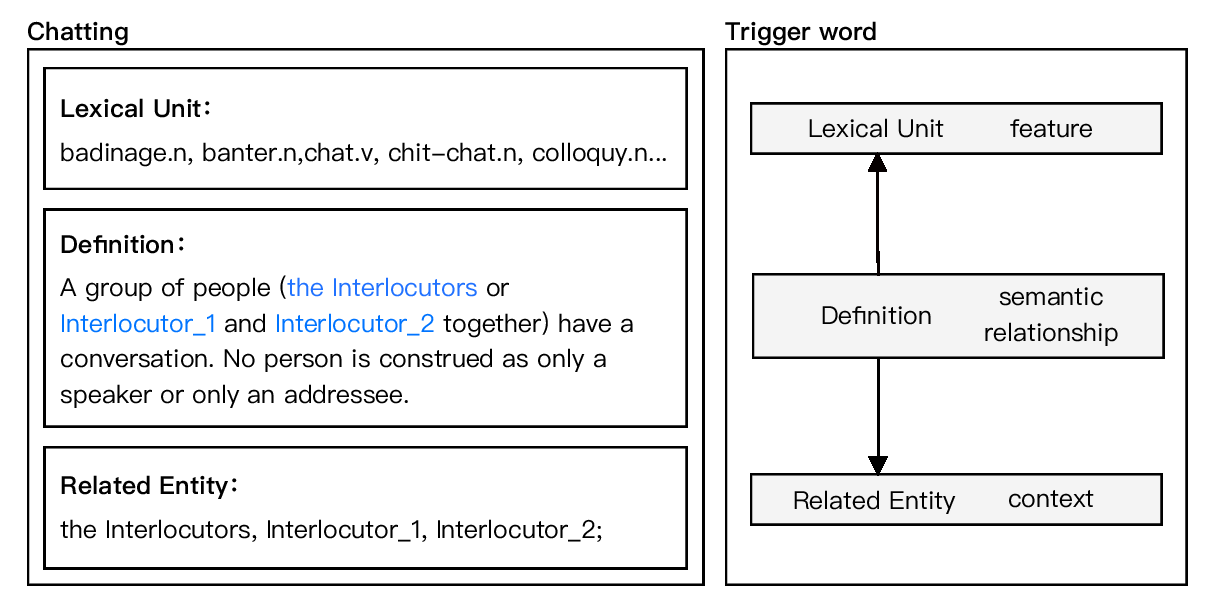}
    \caption{An example of FrameNet.}
    \label{fig:EXTERNAL}
\end{figure*}

A lack of samples causes low training efficiency and aggravates the catastrophic forgetting problem. 
Inspired by Qu \cite{2020-BMLRG-FSRE}, Tong \cite{2020-OPTK-IED}, and Liu \cite{2016-LF-IAED}, 
 we introduce external knowledge to the model to supplement the training samples.

The external semantic knowledge in this case is FrameNet \cite{2002-FrameNet-BFP}, a semantic library of frames proposed by Baker in 2002.
As shown in Fig. \ref{fig:EXTERNAL}, FrameNet consists of a lexical unit and multiple frame elements.
For example, the chatting frame contains the following lexical unit: badinage.n, banter.n, chat.v, chit-chat.n, colloquy.n...; a related entity: the Interlocutors, Interlocutor\_1, Interlocutor\_2; a definition: 'A group of people (the Interlocutors or Interlocutor\_1 and Interlocutor\_2 together) have a conversation. No person is construed as only a speaker or only an addressee.'
A lexical unit represents a feature of a trigger word, an entity represents a context of a trigger word, and a definition represents the semantic relationship between a lexical unit and a trigger word.

We use a knowledge encoder in our model to encode the external knowledge, and a sample encoder to encode the samples.
This guarantees that we can map a sentence and trigger to the same semantic space with the external knowledge. 
We use an adaptive knowledge-enhanced Bayesian framework to ensure the integrity of the knowledge base.
All of the above are shown in the "External Knowledge" module in Fig. \ref{fig:IFSEDK} and \ref{fig:IFSEDKP}.

\noindent\textbf{Knowledge Encoder}
\\
We define a knowledge framework as $F_{t}=\left\{D_{t}, A_{t}, L_{t}\right\}$, where t represents an event class, $D_{t}$ is the definition, $A_{t}$ is the entity, and $L_{t}$ is the lexical unit.
As shown in Fig.\ref{fig:IFSEDK}, $D_{t}^{'}$ and $L_{t}^{'}$ are   generated by preprocessing $D_{t}$ and $L_{t}$ using BERT.
Then, to ensure that all entities mentioned in the definition will be encoded, we convert $A_{t}$ and $D_{t}^{'}$ to $A_{t}^{'}$ by mixed encoding.
Next, we use Attention to get the lexical weight $L_{t}^{*}$ of $L_{t}$ in $D_{t}$ and converge the external knowledge to $A_{t}^{*}$ using $A_{t}^{'}$ and $L_{t}^*$.
Finally, the encoded $L_{t}^{*}$ and $A_{t}^{*}$ are connected.
The information from external knowledge is aggregated into $k_{t}$ by a Feedforward Neural Network.

\noindent\textbf{Sample Encoder}
\\
We define a sample as $X_{s}$.
The sentences in the sample correspond to $D_{t}$, and the trigger words correspond to $L_{t}$ in the knowledge framework.
As shown in Fig. \ref{fig:IFSEDK}, $X_{s}$ is preprocessed using BERT. 
Then the part $X_{s}^{'}$ corresponding to the sentence and the part $T_{s}^{'}$ corresponding to the trigger word are generated.
We use Attention to display the entities $A_{s}^{*}$ by the generated $T_{s}^{'}$ and $X_{s}^{'}$.
Finally, we connect $A_{s}^{*}$ and $T_{s}^{'}$ and use a feedforward network to aggregate the sample information into $s_{t}$.

\noindent\textbf{Adaptive Knowledge-enhanced Posterior}
\\
A learnable knowledge offset ensures that all classes have external knowledge, especially the classes that do not have a match in FrameNet.
This offset can help those classes which do not find in FrameNet match with a similarly class in FrameNet.
The knowledge offset implies the difference between the event class and the knowledge encoding,
\begin{equation}
    \Delta h_{t}=\lambda_{t} \odot\left(s_{t}-k_{t}\right),
\end{equation}
where $\odot\ $ is the element-wise product, $\lambda_{t}$ is generated from the sample encoder and   knowledge encoder, and
\begin{equation}
    \lambda_{t}=\sigma\left(W_{\lambda}\left[s_{t} ; s_{t}-k_{t} ; k_{t}\right]+b_{\lambda}\right),
\end{equation}
where σ is the nonlinear sigmoid function, and $W_{\lambda}$ and $b_{\lambda}$ are trainable parameters.

First, we compute the knowledge prior distribution:
\begin{equation}
p\left(V_{T_s} \mid \mathfrak{F}\right)=\prod_{t \in T_{s}} p\left(v_{t} \mid k_{t}, \Delta h_{t}\right)=\prod_{t \in T_{s}} \mathrm{N}\left(v_{t} \mid k_{t}+\Delta h_{t}, I\right)
\end{equation}
where $V_{T_s}$ is a prototype vector, $\mathfrak{F}$ is the knowledge base, $v_{t}$ is the event class prototype vector, and $\mathrm{N}\left(v_{t} \mid k_{t}+\Delta h_{t}, I\right)$ is a multivariate Gaussian with   mean $k_{t}+\Delta h_{t}$ and covariance I (the identity matrix).

We set each round support set $C_i=S={(a_s,c_s)}$, where $a_s$ and $c_s$ represent the sample and   event class, respectively, of sample $a_s$.
  $A_S=\{{a_s}\}_{s \in S}$ represents the sample in the support S, and 
$C_S=\{{c_s}\}_{s \in S}$ represents the total event class of sample $A_S$.

Then we compute the likelihood for support samples,
\begin{equation}
    p\left(C_{s} \mid A_{S}, V_{T_s}\right)=\prod_{s \in S} p\left(c_{s} \mid a_{s}, V_{T_s}\right)
\end{equation}
and the posterior distribution,
\begin{equation}
    p\left(V_{T_S} \mid A_{S}, C_{S}, \mathfrak{F}\right)=p\left(C_{S} \mid A_{S}, V_{T_S}\right) p\left(V_{T_S} \mid \mathfrak{F}\right)
\end{equation}

We do not use an adaptive knowledge-enhanced posterior in IFSED-KP. 
Instead, we use the output of the knowledge encoder as support and the output of the sample encoder as query in IFSED-KP.

\subsubsection{\textbf{Sample Retention}}
\

A common problem in incremental event detection is catastrophic forgetting, and the few-shot incremental event detection task makes this problem more severe. 
The number of samples in our task is smaller than the number of incremental event detection tasks. 
We use sample retention methods to solve this problem.

Tao argues that true few-shot incremental learning should be trained only on the increments and not on the full data in learning increments \cite{2020-TOPIC-FSCIL}.
However, experimental studies show that sample retention plays an important role in addressing catastrophic forgetting.
To ensure bias for new classes in model learning, we only select one or two samples for retention.
This is shown in the Sample Retention module in Fig. \ref{fig:IFSEDK}.

We define the samples in each new class $C_{i}^{n}$ (the n-th new class in the i-th round) as
\begin{equation}
    X_{s}=\left(S_{i}, L_{i}\right), 1 \leq i \leq k
\end{equation}
where $S_{i}$ represent the sentences in the sample, $L_{i}$ represents the class name corresponding to the sentences in the sample, and $k$ represents the number of samples in the new class, which is usually 5 or 10.

The first step is to select a sample for retention when each new class is learned.
This sample should have the best representation features in the classes the model learned.
We calculate the prototype \cite{2017-PN-FSL} \cite{2018-RC-CPL} to ensure that the selected sample best represents the class.
First, we calculate the prototype of $C_{i}^{n}$ as
\begin{equation}
    \mu_{c}=\frac{1}{k} \sum_{i=1}^{k} S_{i}
\end{equation}
Then we compare each sample in the new class with the prototype.
We calculate the distance between the sample and the prototype, and rank them by distance.
The closer the sample is to the prototype, the more representative it is.
Therefore, we choose the j closest samples as the reserved sample $R_{i}^{n}$ (j is generally taken as 1 or 2 because we are studying few-shot learning).

In the second step, before learning a new class $C_{i}^{n+1}$, where n denotes the nth round, all the retained samples of the previously learned class must be added to the samples to be trained,
\begin{equation}
    R=\left\{R_{1}, \cdots, R_{i-1}, R_{i}^{1}, \cdots R_{i}^{n}\right\}
\end{equation}
where $R_{1}, \cdots, R_{i-1}$ represent the retained samples in the  previous class.

Then the old and new samples $\mathcal{N}=R \cup X_{en}$ (where $X_{en}$ represent the samples of the new class with external knowledge) are learned together.
Since there are more new samples than old samples, the model still focuses more on learning the new class $C_{i}^{n+1}$.
At the same time, the model reviews and recalls the old class.

\subsubsection{\textbf{Model Construction}}
\
\begin{figure*}[!t]
    \centering
    \includegraphics[width=0.89\linewidth]{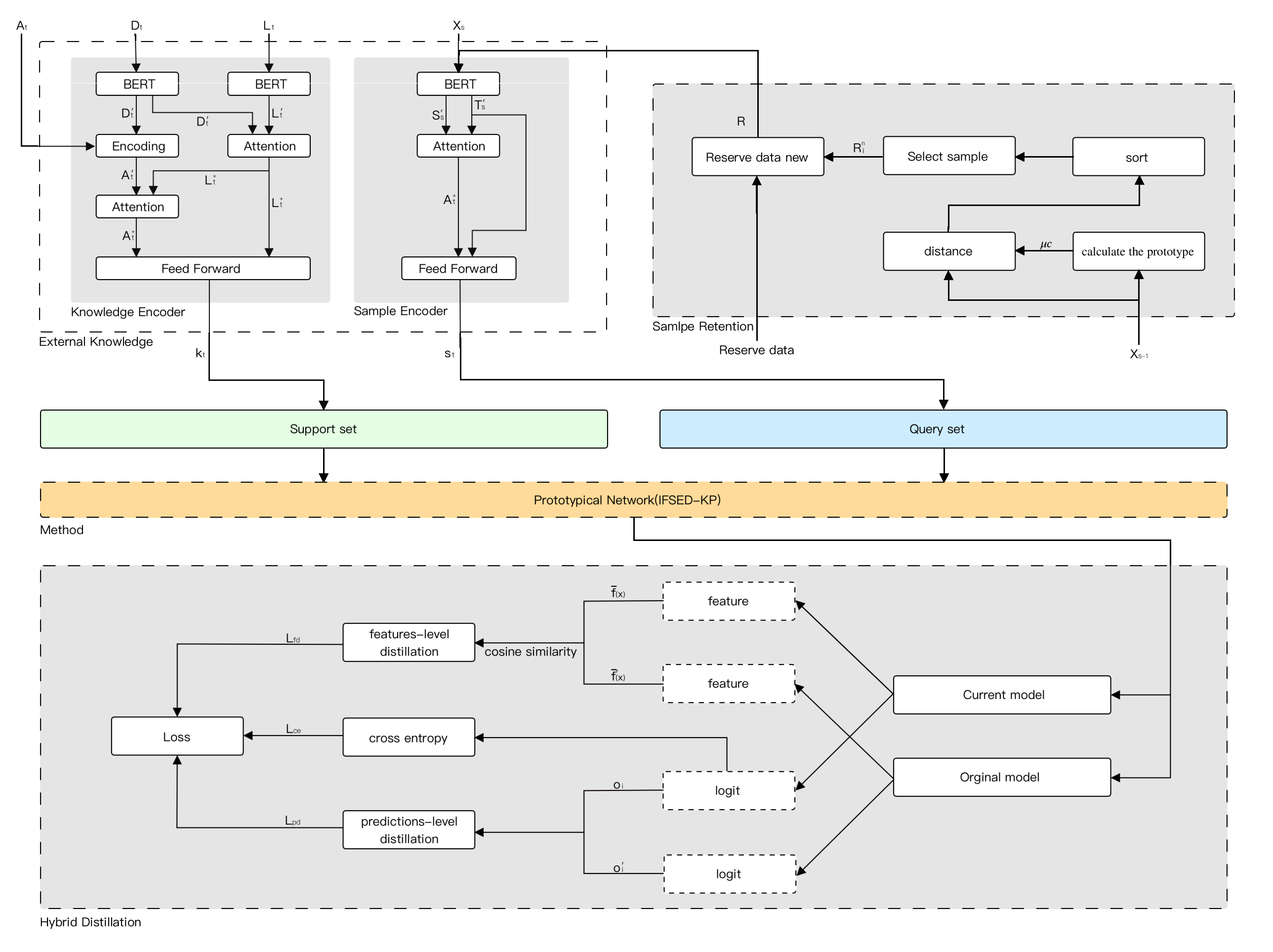}
    \caption{Framework of IFSED-KP model, for which a few-shot incremental event detection model with external knowledge (IFSED-K) is constructed.}
    \label{fig:IFSEDKP}
\end{figure*}

\noindent\textbf{IFSED-K}
\\ 
As shown in Fig. \ref{fig:IFSEDK}, we use the general classifier approach for model construction, for which a few-shot incremental event detection model with external knowledge (IFSED-K) is constructed.
The model uses the external knowledge references mentioned above, aligns the external knowledge with the training data, fuses the two, and uses a classifier for learning.
Distillation is then performed using a hybrid distillation method (described below). 
This can ensure that new classes are adequately learned while reducing instances of catastrophic forgetting.
The model learns N classes at a time, where each class contains K samples.
After each learning round, samples to be retained are selected and added to the next round for model training.

\noindent\textbf{IFSED-KP}
\\
As shown in Fig. \ref{fig:IFSEDKP}, we also use a meta-learning approach for model construction, for which a few-shot incremental event detection model with external knowledge and prototype network (IFSED-KP) is constructed.
The prototype network is considered one of the better solutions to few-shot learning at this stage.
We incorporate external knowledge to improve its learning efficiency.
This construction method is different from the general meta-learning model.
To fit the problem of limited sample size in real life, we fuse the definition and vocabulary units from external knowledge as support and use the training data as query for input, without using the dev-dataset.
To diminish catastrophic forgetting, sample retention and hybrid distillation are incorporated in this model.

\subsubsection{\textbf{Hybrid Distillation}}
\

One cause of the catastrophic forgetting problem is class imbalance, i.e., after sample retention, the new class has a larger sample, and the old class has a smaller sample, at which time the model will be biased toward the old class, leading to severe forgetting of the previous knowledge \cite{2019-LSIL-IL} \cite{2019-REB-IL}.
Knowledge distillation was proposed to transfer knowledge from one network to another \cite{2015-DK-NN}.

\noindent\textbf{Cross-entropy loss function}
\\
We use cross-entropy loss, as is common in classification tasks, as the main loss function.
Because the old and new data are learned simultaneously, this is calculated as
\begin{equation}
    \mathcal{L}_{c e}=-\frac{1}{|\mathcal{N}|} \sum_{X_{e} \in \mathcal{N}} \sum_{x \in X_{e}} y_{i} \log p  
\end{equation}
where $y_{i}$ is the one-hot ground-truth label for token $x$, and $p$ is the corresponding class probability obtained by softmax.

\noindent\textbf{Feature-level Distillation}
\\
Feature-level Distillation is used mainly for scenarios where the features extracted by the current model are not very different from those extracted by the previous model.
Using a feature extractor in this case will effectively help the model to retain the features of knowledge.
Therefore, we use a Feature-level Distillation loss function to help the model with knowledge retention:
\begin{equation}
    \mathcal{L}_{f d}=\frac{1}{|\mathcal{N}|} \sum_{X_{e} \in \mathcal{N}} \sum_{x \in X_{e}} 1-<\bar{f}^{\prime}(x), \bar{f}(x)>
\end{equation}
where $\bar{f}^{\prime}(x)$ and $\bar{f}(x)$ are l2-normalized features extracted from the original and current model, respectively, and $<\bar{f}^{\prime}(x), \bar{f}(x)>$ measures the cosine similarity between two normalized feature vectors.

\noindent\textbf{Predictions-level Distillation}
\\
We also use Feature-level Distillation, which helps preserve previously learned knowledge by encouraging the current model's predictions for new classes to match the class names of previous models.
The Prediction-level Distillation loss function is
\begin{equation}
    \mathcal{L}_{p d}=-\frac{1}{|\mathcal{N}|} \sum_{X_{e} \in \mathcal{N}} \sum_{x \in X_{e}} \sum_{i=1}^{m} \mathcal{T}_{i}^{\prime} \log \left(\mathcal{T}_{i}\right)
\end{equation}
where
\begin{equation}
    \mathcal{T}_{i}^{\prime}=\frac{e^{o_{i}^{\prime} / T}}{\sum_{j=1}^{m} e^{o_{i}^{\prime} / T}}
\end{equation}
\begin{equation}
    \mathcal{T}_{i}=\frac{e^{o_{i} / T}}{\sum_{j=1}^{m} e^{o_{i} / T}}
\end{equation}
where $T$ is the temperature scalar, which is usually set to be greater than 1 to increase the weights of small values; $m$ is the number of old classes observed by the model when the new class arrives; $o_{i}^{'}$ is the output logit of the old model; and $o_{i}$ is the output logit of the new model.

The total loss is
\begin{equation}
    \mathcal{L}=\alpha \mathcal{L}_{f d}+\beta \mathcal{L}_{p d}+\gamma \mathcal{L}_{c e}
\label{loss_weight}
\end{equation}
where α and β are adjustment coefficients. 
If α, β, and γ are larger, the model will give more weight to the old class, and if α and β are smaller, the model will give more weight to the new class. 
The values of α, β, and γ will be experimentally demonstrated, and an optimal value will be selected for application in the model.

\section{Experiments}
\label{sec:experiments}
We describe the dataset and experimental settings, evaluate our model in different ways, analyze the results of the model, and find the influence factors in tests.  

\subsection{Dataset}
FewEvent is a few-shot event detection dataset constructed by Deng \cite{2020-DMBPN-FSED} in 2020.
It is the largest dataset in few-shot event detection learning, with 100 event types and 70,852 samples.

\begin{figure*}[!t]
    \centering
    \includegraphics[width=0.6\linewidth]{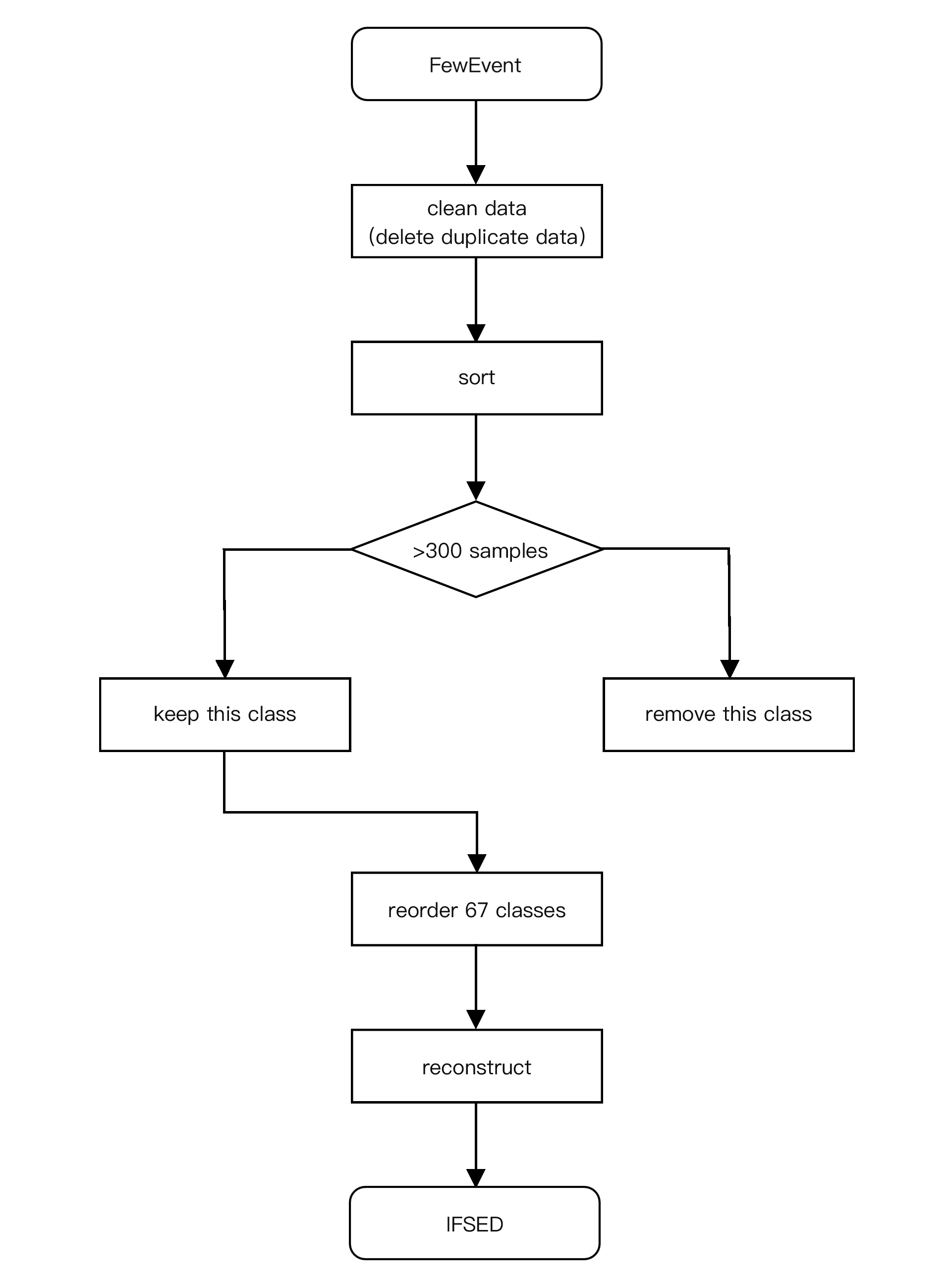}
    \caption{This flowchart shows how we reconstruct the FewEvent dataset into our IFSED dataset.}
    \label{fig:Flowchart}
\end{figure*}

As shown in Fig \ref{fig:Flowchart}, we adapted the FewEvent dataset IFSED to accommodate the event detection task.
The data were cleaned. 
Duplicate samples were deleted to avoid duplication in the training, development, and testing sets.
Each class was sorted by number, and 67 classes with more than 300 samples were selected. 
The other classes lacked sufficient data for training after cleaning.
Finally, we reordered them and reconstructed the dataset according to the specifications in Table \ref{table:dataset}.

\begin{figure*}[!t]
    \centering
    \includegraphics[width=0.89\linewidth]{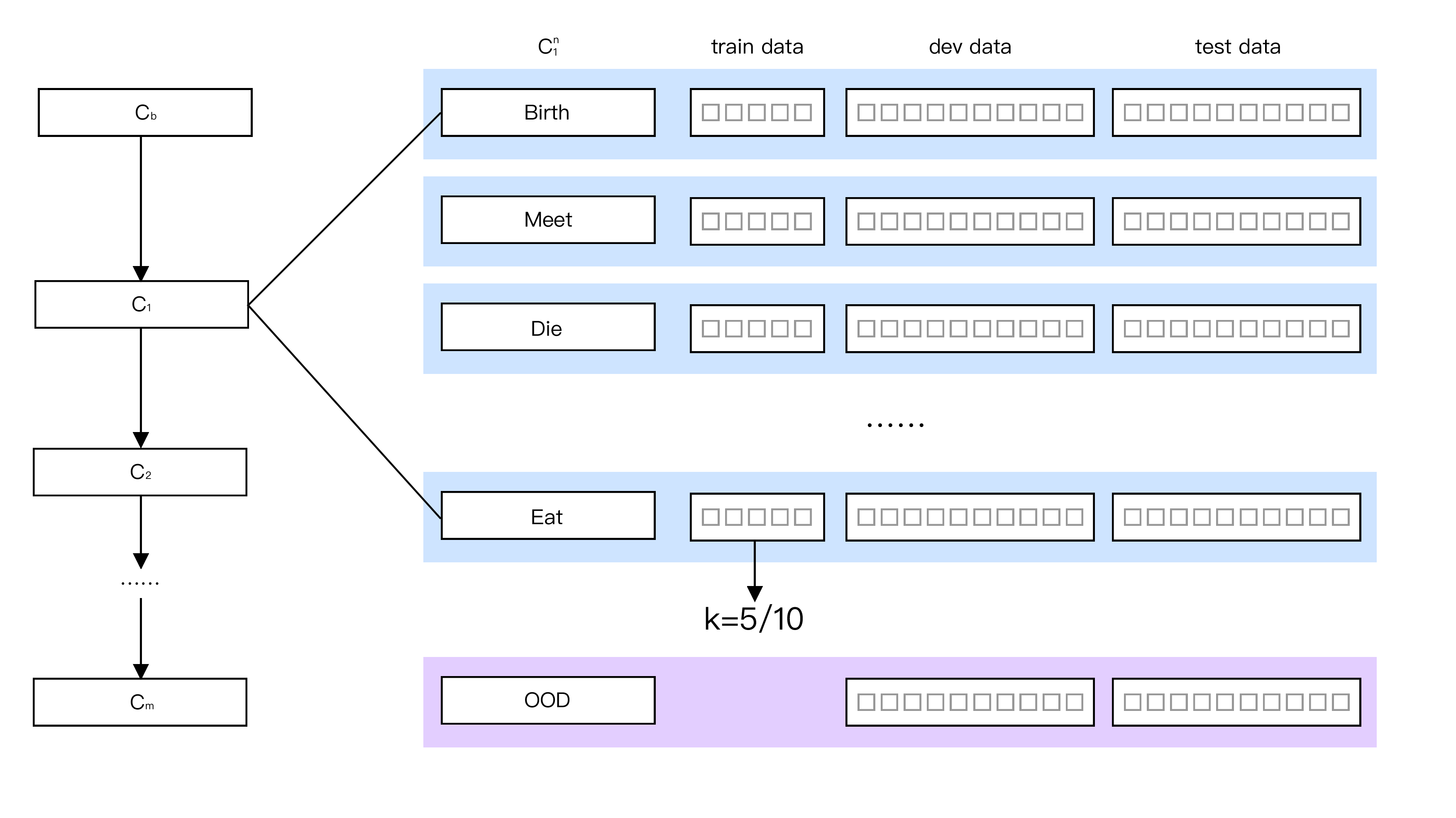}
    \caption{Base classes and new classes have contained training, development, and test sets. OOD classes only have development set and test sets.}
    \label{fig:construction}
\end{figure*}

As shown in Fig \ref{fig:construction}, we constructed 5-way-5-shot and 10-way-10-shot datasets, each including base classes, new classes, and OOD classes.
Base classes and new classes have contained training, development, and test sets. 
OOD classes only have development set and test sets.
Each of the 10 base classes in the training set had 100 samples, and the development and test sets both had 50 samples.
This new kind of class consists of 5 rounds, and each round contains N classes that need to be learned. 
Each new class had K samples in the training set, and 10 samples each in the development and test sets.
The OOD kind consisted of seven classes and lacked a training set. 
Each OOD class had 15 samples in both the development and test sets.
The dataset was rebuilt to conform to the data requirement of few-shot incremental learning, and no overlapping samples existed.
An overview of the dataset is shown in Table \ref{table:dataset}.

We did not use the base data and development set in experiments, but we still created it and describe it in this paper, as we hope this new dataset will be used in further research.

\begin{table}[]
\begin{tabular}{|c|c|c|c|c|c|c|c|c|}
		\hline

            & \multicolumn{4}{c|}{5-way-5-shot}                                            & \multicolumn{4}{c|}{10-way-10-shot} \\
            & \multicolumn{1}{c|}{\#class} & \multicolumn{1}{c|}{\#train} & \#dev & \#test & \#class & \#train & \#dev & \#test \\ \hline
$c_{b}$     & \multicolumn{1}{c|}{10}      & \multicolumn{1}{c|}{1000}    & 500   & 500    & 10      & 1000    & 500   & 500    \\
$c_{1}$     & \multicolumn{1}{c|}{5}       & \multicolumn{1}{c|}{25}      & 50    & 50     & 10      & 100     & 100   & 100    \\
$c_{2}$     & \multicolumn{1}{c|}{5}       & \multicolumn{1}{c|}{25}      & 50    & 50     & 10      & 100     & 100   & 100    \\
$c_{3}$     & \multicolumn{1}{c|}{5}       & \multicolumn{1}{c|}{25}      & 50    & 50     & 10      & 100     & 100   & 100    \\
$c_{4}$     & \multicolumn{1}{c|}{5}       & \multicolumn{1}{c|}{25}      & 50    & 50     & 10      & 100     & 100   & 100    \\
$c_{5}$     & \multicolumn{1}{c|}{5}       & \multicolumn{1}{c|}{25}      & 50    & 50     & 5       & 100     & 100   & 100    \\
OOD     & \multicolumn{1}{c|}{7}       & \multicolumn{1}{c|}{-}       & 105   & 105    & 7       & -       & 105   & 105    \\
\hline

\end{tabular}
\caption{Statistics of datasets for 5-way-5-shot and 10-way-10-shot. Development set is not used in this paper.}
\label{table:dataset}
\end{table}

\subsection{Experimental settings}
\subsubsection{\textbf{Baseline}}
\ 

As a newly proposed method, the task of few-shot incremental event detection has no model. 
We used the FINETUNE, KCN, and AKE models as baselines.

\noindent\textbf{FINETUNE}
\

This is a simple approach to fine-tune the model as new classes arrive. 
We used the state-of-the-art BERT as a pretrained model.

\noindent\textbf{KCN}
\ 

Originally used for incremental event detection.
KCN uses a prototype enhanced retrospection and hierarchical distillation approach to solve the catastrophic forgetting problem.
We changed the input of KCN to fit our few-shot incremental event detection task.

\noindent\textbf{AKE}
\ 

 Originally used for the few-shot event detection task, AKE uses Adaptive Knowledge-Enhanced Bayesian Meta-Learning to address the problem of insufficient sample sizes. To fit our task, we removed the adaptive Bayesian network and used external knowledge as support and training data as a query.



\subsubsection{\textbf{Implementation and setting}}
\ 

The F1-score has been used as an evaluation metric in previous work \cite{2015-DMPCN-EE} \cite{2018-ABGIA-EE} \cite{2020-EGCN-EE}.
The parameters of IFSED-K were set as follows: epoch was 50; $\alpha$ , $\beta$, and $\gamma$ in formula \ref{loss_weight} were 0.1, 0.1, and 0.5 respectively; and the number of record samples was 1. 
The parameters of IFSED-KP were as follows: epoch was 500; $\alpha$ , $\beta$ , and $\gamma$ were 0.01, 0.5, and 0.7, respectively; and the number of record samples was 1.

\subsection{Results}
We explore three questions:
1) Can our system obtain better performance on each round?
2) Do different sample sizes and numbers of new classes affect the results?
3) Do different numbers of retained samples affect the results?

\subsubsection{\textbf{Few-shot Incremental Event Detection Learning}}
\ 

\noindent\textbf{Main experiment}
\ 

Tables \ref{table:main} and \ref{table:main_forgetting_rate} compare the results of FINETUNE, KCN, and AKE with those of IFSED-KP and IFSED-K, using the same five rounds of the 5-way-5-shot dataset as input.
Table \ref{table:main} shows the results of each learning, and Table \ref{table:main_forgetting_rate} shows the forgetting rates calculated from these. 
The forgetting rate is
\begin{equation}
    Rate_{forgetting} = \frac{\sum (p_n-p_{n-1})/p_n}{4}  
\end{equation}
where $p_n$ is the average F1-score of classes that have been learned before the n-th round.

\begin{table}[]
\begin{tabular}{|cl|llllll|}
		\hline

                                              & Model    & $c_{1}$        & $c_{2}$        & $c_{3}$        & $c_{4}$        & $c_{5}$        & OOD\\ \hline
\multicolumn{1}{|c|}{\multirow{5}{*}{$ c_{1}$}} & FINETUNE & \textbf{51.40} &                &                &                &              &\ \ \ - \\
\multicolumn{1}{|c|}{}                         & KCN      & 48.50          &                &                &                &               &\ \ \ -\\
\multicolumn{1}{|c|}{}                         & AKE      & 46.95          &                &                &                &               & 16.57 \\
\multicolumn{1}{|c|}{}                         & IFSED-KP & 48.19          &                &                &                &               & \textbf{16.99} \\
\multicolumn{1}{|c|}{}                         & IFSED-K  & 49.50          &                &                &                &               &\ \ \ -\\\hline
\multicolumn{1}{|c|}{\multirow{5}{*}{$ c_{2}$}} & FINETUNE & 4.50           & \textbf{52.20} &                &                &              &\ \ \ -\\
\multicolumn{1}{|c|}{}                         & KCN      & 37.05          & 49.10          &                &                &               &\ \ \ -\\
\multicolumn{1}{|c|}{}                         & AKE      & 26.59          & 27.20          &                &                &               & 17.47\\
\multicolumn{1}{|c|}{}                         & IFSED-KP & 26.79          & 27.42          &                &                &               & \textbf{22.78} \\
\multicolumn{1}{|c|}{}                         & IFSED-K  & \textbf{39.90} & 49.35          &                &                &               &\ \ \ -\\\hline
\multicolumn{1}{|c|}{\multirow{5}{*}{$c_{3}$}} & FINETUNE & 0.00           & 0.00           & \textbf{67.55} &                &               &\ \ \ -\\
\multicolumn{1}{|c|}{}                         & KCN      & 19.15          & 24.85          & 65.00          &                &               &\ \ \ -\\
\multicolumn{1}{|c|}{}                         & AKE      & 19.10          & \textbf{28.50} & 53.00          &                &               & 9.15\\
\multicolumn{1}{|c|}{}                         & IFSED-KP & 19.55          & 27.06          & 31.95          &                &               & \textbf{20.16} \\
\multicolumn{1}{|c|}{}                         & IFSED-K  & \textbf{24.70} & 25.55          & 63.30          &                &               &\ \ \ -\\\hline
\multicolumn{1}{|c|}{\multirow{5}{*}{$c_{4}$}} & FINETUNE & 0.00           & 0.00           & 8.05           & \textbf{79.70} &               &\ \ \ -\\
\multicolumn{1}{|c|}{}                         & KCN      & 23.40          & 24.50          & 49.75          & 76.80          &               &\ \ \ -\\
\multicolumn{1}{|c|}{}                         & AKE      & 25.20          & 27.80          & 45.80          & 50.80          &                & 13.91\\
\multicolumn{1}{|c|}{}                         & IFSED-KP & 25.40          & 27.87          & 37.68          & 51.40          &                & \textbf{30.62}\\
\multicolumn{1}{|c|}{}                         & IFSED-K  & \textbf{27.80} & \textbf{28.80} & \textbf{50.75} & 73.55          &                &\ \ \ -\\\hline
\multicolumn{1}{|c|}{\multirow{5}{*}{$c_{5}$}} & FINETUNE & 0.00           & 0.00           & 1.45           & 21.35          & \textbf{76.10} &\ \ \ -\\
\multicolumn{1}{|c|}{}                   & KCN      & 18.65          & 21.40          & 50.60          & 65.15          & 75.55          &\ \ \ -\\
\multicolumn{1}{|c|}{}                   & AKE      & 19.15          & 25.10          & 38.25          & 40.60          & 56.90          & 15.22\\
\multicolumn{1}{|c|}{}                   & IFSED-KP & 20.10          & \textbf{26.07} & 32.66          & 42.00          & 50.05          & \textbf{24.13}\\
\multicolumn{1}{|c|}{}                   & IFSED-K  & \textbf{22.05} & 25.15          & \textbf{50.65} & \textbf{65.65} & 71.85         &\ \ \ -\\

\hline

\end{tabular}
\caption{Comparison of model performance under 5-way-5-shot without base classes. Horizontal axis: groups of 
test classes ($C_1,...,C_5$) and OOD classes ($OOD$);
vertical axis: timeline of incremental learning over new rounds of novel classes.}
\label{table:main}
\end{table}

\begin{table}[]
\begin{tabular}{|l|l|}
\hline
Model   & Forgetting rate \\ \hline
FINETUNE & 70.33\%         \\
KCN      & 26.05\%         \\
AKE      & 19.92\%         \\
IFSED-KP & \textbf{16.42}\%         \\
IFSED-K  & 22.54\%  \\
\hline
\end{tabular}
\caption{Comparison of forgetting rates under 5-way-5-shot without base classes}
\label{table:main_forgetting_rate}
\end{table}

According to the results, we found that the abilities to preserve the knowledge of KCN and AKE had been improved and were obviously better than those of the FINETUNE model.
In new class learning, they obviously performed worse than the FINETUNE model, but it can be clearly seen that they had a higher recognition ability of previous classes.

Similarly, IFSED-KP and IFSED-K learned new classes normally because they paid more attention to previous classes, and both showed good results for previous class detection and well solved the catastrophic forgetting problem with few-shot incremental event detection.
IFSED-K had good stability and better scoring ability than other models. 

IFSED-KP outperformed AKE in the detection of previous classes; both use meta-learning methods.
IFSED-KP also outperformed AKE on OOD data.
The meta-learning models AKE and IFSED-KP had a lower forgetting rate.
The meta-learning model is well suited to solve the few-shot incremental event detection task.
	
\noindent\textbf{Ablation experiments}
\ 

\begin{table}[]
\begin{tabular}{|l|lllll|}
\hline
Model  & $New$            & $p_1$            & $p_2$            & $p_3$            & $p_4$            \\\hline
IFSED-K & 61.51          & 45.46 & \textbf{34.72} & \textbf{26.48} & \textbf{22.05} \\
w/o EK  & \textbf{62.99} & 44.20          & 31.42          & 22.40          & 18.65          \\
w/o ML  & 61.57          & 44.15          & 31.66          & 22.01          & 18.50          \\
w/o $L_{fd}$       & 62.40         & 39.30          & 21.59          & 13.06          & 7.69 \\
w/o $L_{pd}$       & 62.80          & \textbf{45.66}         & 32.71          & 20.96          & 14.81 \\
w/o PS  & 62.32          & 8.65           & 0.43           & 0.00           & 0.00      \\
\hline
\end{tabular}
\caption{Ablation experiments: IFSED-K}
\label{table:Ablation experiments IFSED-K}

\end{table}

\begin{table}[]
\begin{tabular}{|l|llllll|}
\hline
Model           & $New$            & $p_1$            & $p_2$            & $p_3$            & $p_4$             & $OOD$\\\hline
IFSED-KP         & 41.80          & \textbf{33.38} & 26.96       & \textbf{25.76}         &21.10         & \textbf{22.94}\\
w/o ML and PS    & \textbf{48.97}          & 33.35          & 26.62          & 25.15          & 19.15         & 14.46\\
w/o $L_{fd}$       & 35.87          & 31.34          & \textbf{27.19}          & 24.61          & \textbf{24.89}         & 18.99\\
w/o $L_{pd}$       & 36.51          & 29.17          & 26.19          & 23.67          & 23.43         & 15.85\\
\hline
\end{tabular}
\caption{Ablation experiments: IFSED-KP}
\label{table:Ablation experiments IFSED-KP}

\end{table}

Ablation experiments further demonstrated the effectiveness of adding external knowledge, sample retention, and hybrid distillation, as shown in Tables \ref{table:Ablation experiments IFSED-K} and \ref{table:Ablation experiments IFSED-KP}, where
EK stands for External Knowledge, ML for Mixture Loss, and PS for Prototype-based Selection.
Furthermore, the data from the current and previous rounds are presented to show the validity of the model.

In the ablation experiment of IFSED-K, the F1-score decreased after removing external knowledge.
The more rounds, the more significant the drop in score.
This shows that external knowledge can effectively help the model to learn.
It also has a relieving effect on catastrophic forgetting.
After removing the hybrid loss function, it is observed that the learning ability of the model for new classes decreases substantially, and likewise for previous classes.
This shows that the addition of hybrid distillation plays an important role in both the bias of new classes and the retention of samples and alleviates the problem of class imbalance.
When sample retention is removed, the learning ability of a new class is improved, but that of the previous class is substantially weakened.
Sample retention plays an important role in the general approach of few-shot incremental event detection.

In ablation experiments of IFSED-KP, the model used external knowledge in support and training samples in queries. 
The removal of external knowledge does not conform to the basic structure of the meta-learning model.
Therefore, a control group with external knowledge removed was not set up in the ablation experiments.
After removing the sample retention and mixture distillation, the learning ability of a new class was improved. 
However, at the same time, there was a certain decrease in learning ability for previous classes and OOD data.
In this way, sample retention and hybrid distillation play a role in few-shot incremental event detection models using meta-learning methods.
However, comparing the two models, it can be found that sample retention in the meta-learning method does not play a major role compared with the general method, perhaps because the meta-learning approach focuses on how to discriminate different classes rather than learning a particular class.

\subsubsection{\textbf{Effects of different ways and shots}}
\ 

\begin{figure*}
    \centering
    \subfigure[IFSED-K]
    {
    \centering
    \includegraphics[scale=0.3]{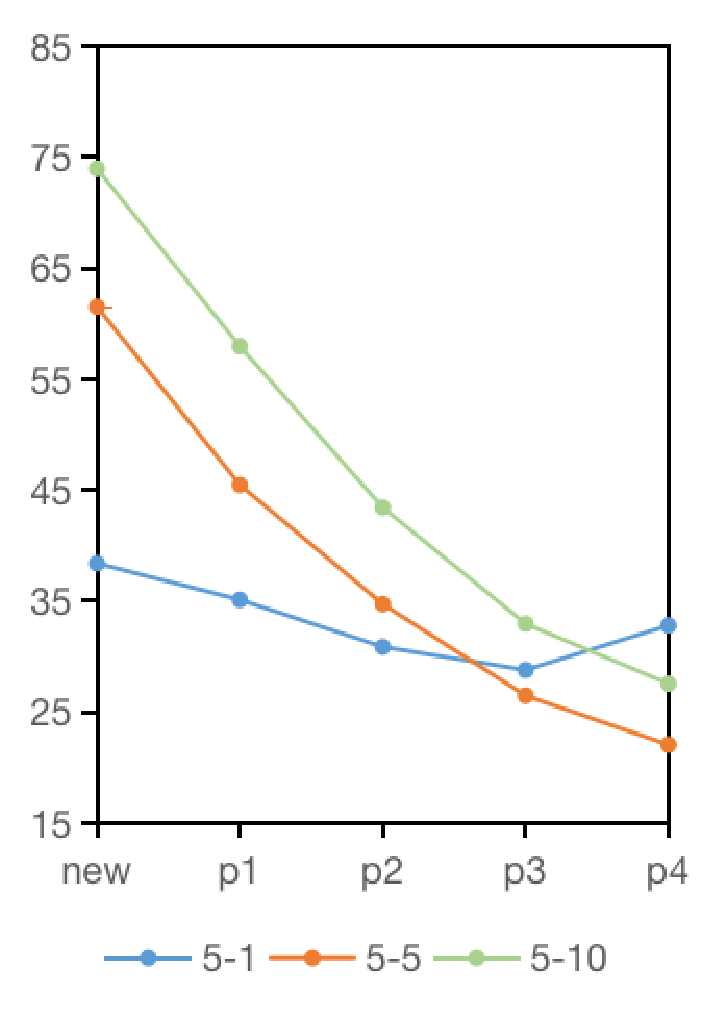}
    \label{fig:DS-IFSEDK}
    }
    \subfigure[IFSED-KP]
    {
    \centering
    \includegraphics[scale=0.3]{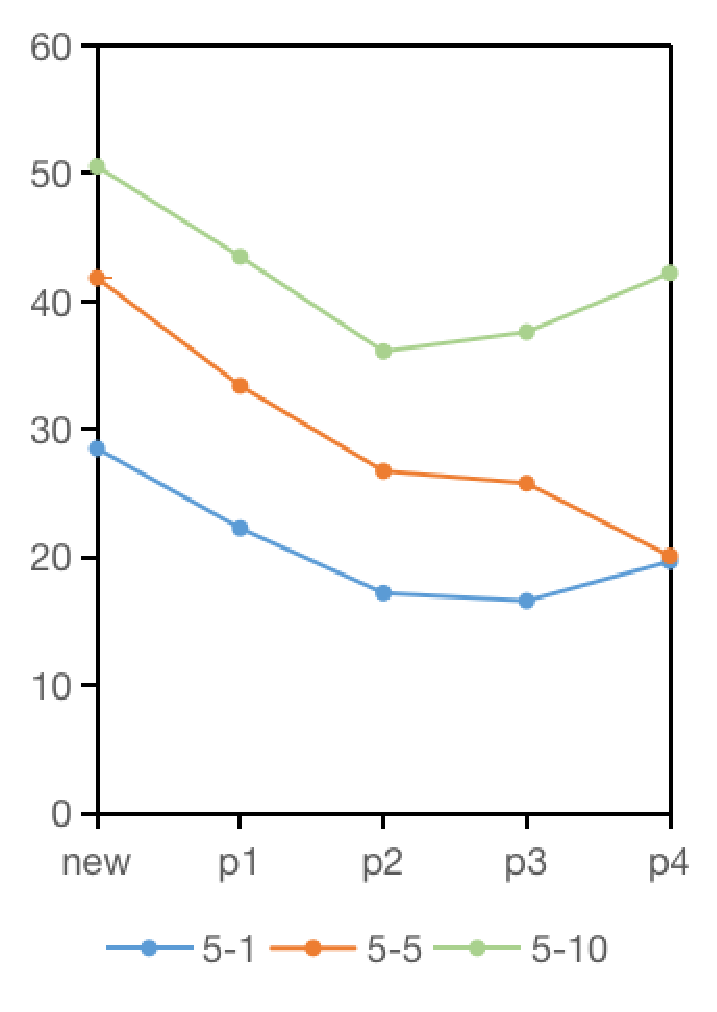}
    \label{fig:DS-IFSEDKP}
    }
    \caption{Effects of different numbers of shots}
\end{figure*}

\begin{figure*}
    \centering
    \subfigure[IFSED-K]
    {
    \centering
    \includegraphics[scale=0.3]{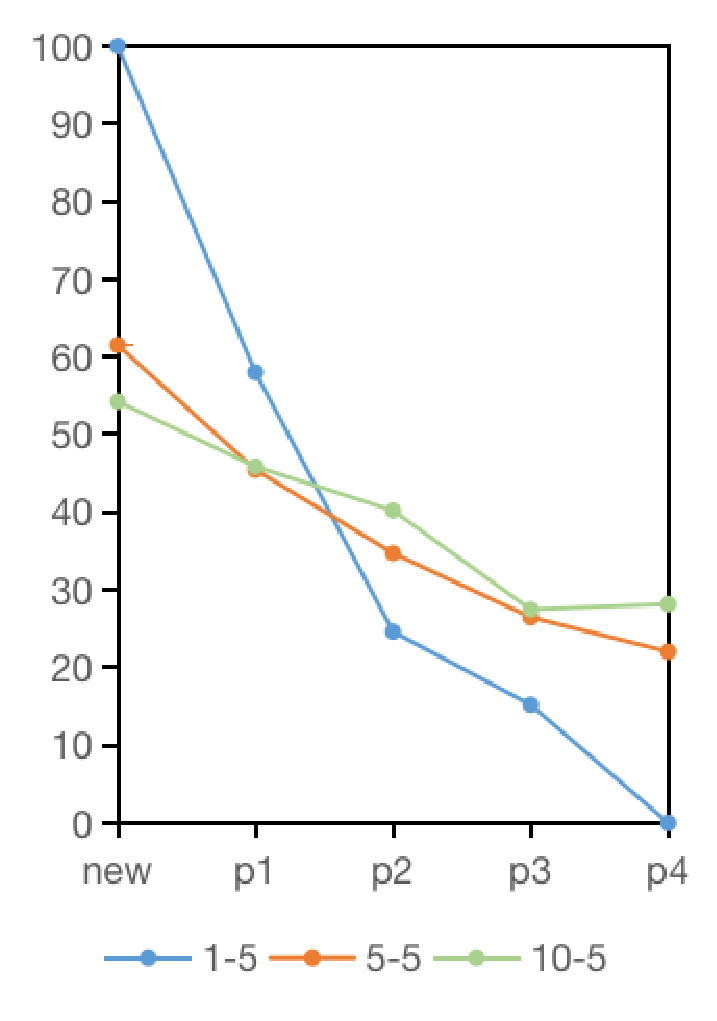}
    \label{fig:DW-IFSEDK}
    }
    \subfigure[IFSED-KP]
    {
    \centering
    \includegraphics[scale=0.3]{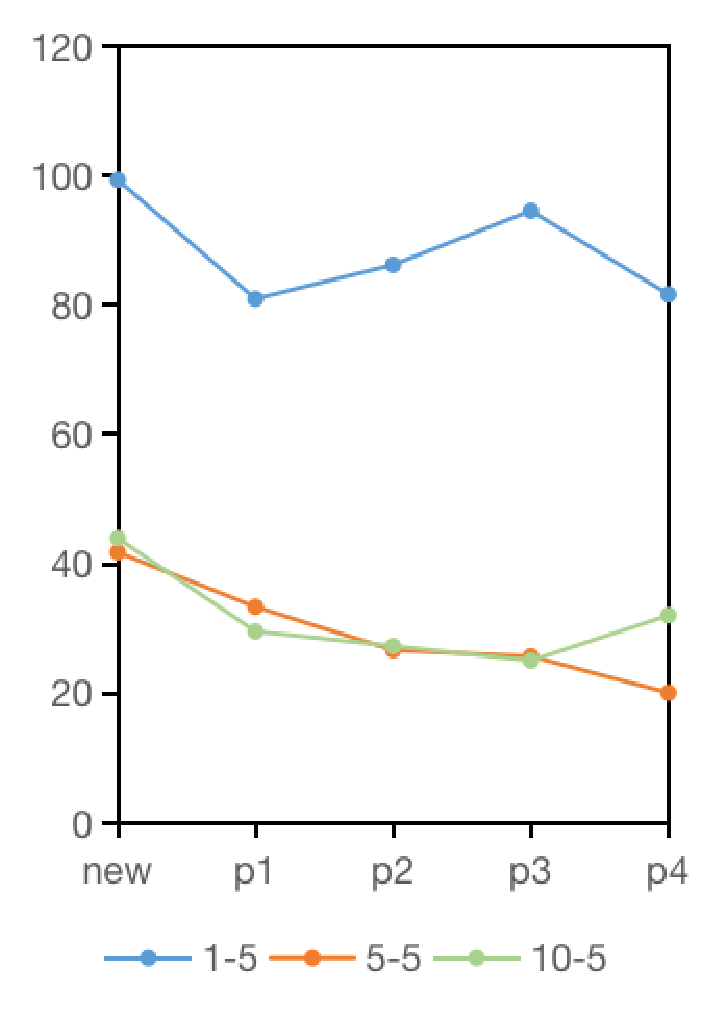}
    \label{fig:DW-IFSEDKP}
    }
    \caption{Effects of different numbers of ways}
\end{figure*}

We conducted 5-way-1-shot, 5-way-5-shot, and 5-way-10-shot experiments and 1-way-5-shot, 5-way-5-shot, and 10-way-5-shot experiments on IFSED-K and IFSED-KP, respectively, using the statistical method described above, where the average of the current round and previous rounds was calculated to show a model's ability in different rounds.

As shown in Fig. \ref{fig:DS-IFSEDK} and \ref{fig:DS-IFSEDKP}, we can find the effect of different numbers of retained samples. 
Since only the data from the first round of learning were available, their representation was low in the previous four rounds ($p_4$).
However, it can be seen that the curve trends of both IFSED-K and IFSED-KP were roughly consistent. 
In experiments with different sample sizes, 5-way-10-shot performed better in both IFSED-K and IFSED-KP.

Experimental results with 5-way-10-shot as input were better than with 5-way-1-shot and 5-way-5-shot in terms of new class learning and old class retention, showing that an enhanced sample size   better guides the learning of the model.
It is inferred that the improvement of sample size has a huge effect on the model in the task we defined.
However, how to further improve the accuracy with few-shot samples is a problem to be solved.

As shown in Fig. \ref{fig:DW-IFSEDK} and \ref{fig:DW-IFSEDKP}, we can determine the effect of different numbers of new classes as input, and it can be seen that the 5-way-5-shot and 10-way-5-shot line is mostly match. 
While 1-way-5-shot had a higher forgetting rate in IFSED-K, it performed better in IFSED-KP, perhaps because the total sample size was small in IFSED-K. 
Therefore, the current parameters are more biased toward new classes, which leads to catastrophic forgetting.
By contrast, in IFSED-KP, the meta-learning model focuses on learning how to classify the class.
Therefore, the reduction in the total number of classes enables the model to better distinguish classes. 
When the number of classes increased beyond 5, the effects of the above advantages and disadvantages were substantially diminished, leading to more consistent performance of the model.

\subsubsection{\textbf{Effects of   numbers of retained samples}}
\ 
\begin{figure*}
    \centering
    \subfigure[5-way-5-shot]
    {
    \centering
    \includegraphics[scale=0.3]{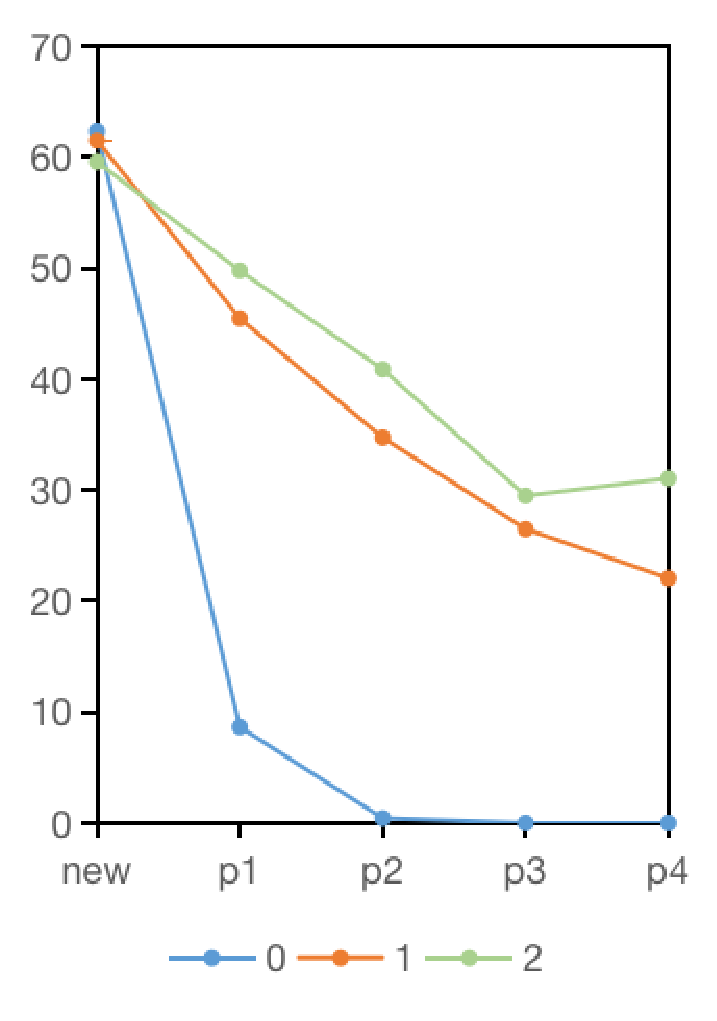}
    \label{fig:5-way-5-shot}
    }
    \subfigure[10-way-10-shot]
    {
    \centering
    \includegraphics[scale=0.3]{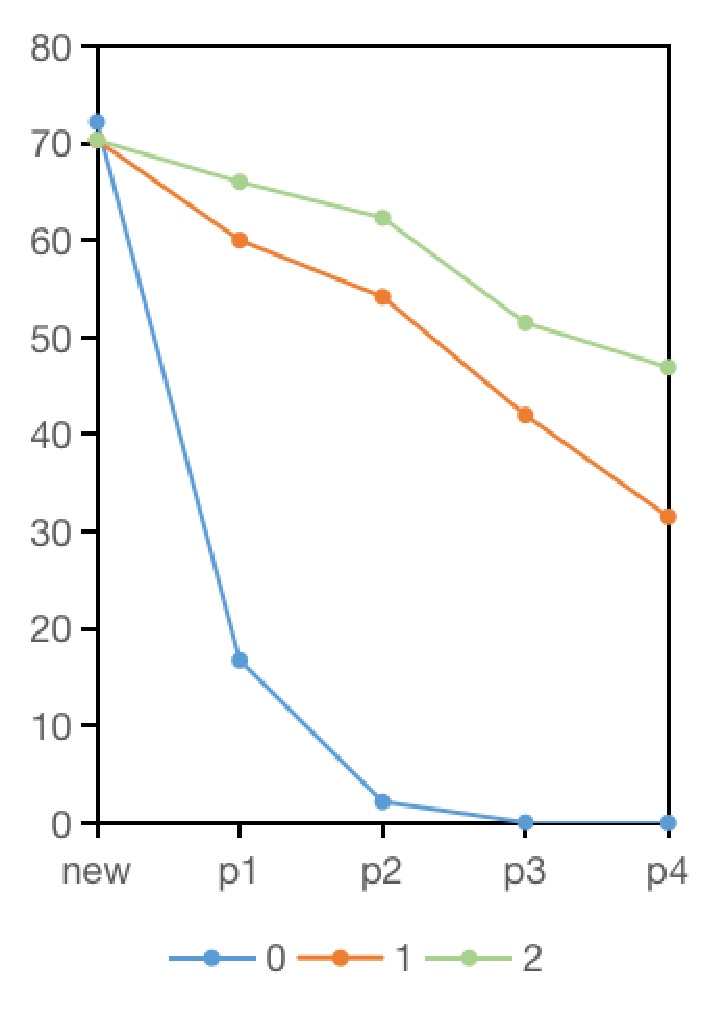}
    \label{fig:10-way-10-shot}
    }
    \caption{Effect of different numbers of retained samples}
\end{figure*}

We conducted experiments using 5-way-5-shot and 10-way-10-shot in IFSED-K to retain zero, one, and two samples, still using the method described above, calculating the average of the current and   previous rounds to show the model's ability in different rounds.
As shown in Fig. \ref{fig:5-way-5-shot} and  \ref{fig:10-way-10-shot}, the more samples retained in training, the better the performance in detecting the previous classes.
The difference between retaining one sample and not retaining samples is obvious, as the memory effect of classes is substantially improved, further demonstrating the effectiveness of introducing sample retention in this model.

\section{Related Work}
\label{sec:related-work}

\subsection{Incremental learning}
Incremental learning was defined in a paper like this: it can learn additional information from new data; it does not require access to previously used data; it retains a substantial majority of the previously acquired knowledge when learning new information; it can learn instances of previously unseen classes; it is versatile enough to be used with a variety of classification algorithms \cite{2002-ILA}. 
However, certain models do not follow this definition strictly and retain the previous data during training, because they recognize the significant contribution of previous data to improve the validity of the model.

A widely proven and studied problem in current research on incremental learning is the problem of catastrophic forgetting due to the tendency for knowledge of the previously learned task(s) (e.g., task A) to be abruptly lost as information relevant to the current task (e.g., task B) is incorporated. 
This phenomenon occurs specifically when the network is trained sequentially on multiple tasks because the weights in the network that are important for task A are changed to  meet the objectives of task B. \cite{2016-OCF}

To address this problem, incremental learning has evolved various approaches, which are classified into regularization-based, replay-based, and parameter isolation-based methods. 
In this section, we focus on two related models for incremental event detection.

Cao first proposed the task of incremental event detection and KCN \cite{2020-KCN-IED} in his paper.
KCN adopts prototype-enhanced retrospection and hierarchical distillation to mitigate the adverse effects of semantic ambiguity and class imbalance, respectively.

KPGNN \cite{2021-KPGNN-IED} transforms complex social messages into unified social graphs using Heterogeneous GNNs.
It leverages the inductive learning ability of GNNs to detect events efficiently and extends its knowledge from previously unseen data. 
After adding new data, the system will detect automate. 
It will extend the original social graph and remove outdated messages periodically to maintain an up-to-date social graph structure.

\subsection{Few-shot learning}
Few-Shot is defined as FSL (Few-Shot Learning) as a machine learning problem which contains only a limited number of labeled examples. \cite{2020-surveyFSL}
The training set of Few-shot contains multiple categories with multiple samples per category.
In the training phase, N categories are drawn in the training set, with K samples per category.
First, a meta-task is constructed as the input to the support set of the model.
And then a batch of the remaining data from it is extracted as the prediction object of the model.
The model is required to learn how to distinguish between the categories from the individual data, and such a task is called the N-way-K-shot problem.

At present, there are two main approaches of few-shot learning, one is generative model-based approaches. 
The other one is discriminate model-based approaches \cite{2020-FSLearning-survey}.
The generative model-based approaches mainly use non-deep learning methods.
The discriminate model-based approaches mainly consist of a feature extractor and a predictor.

Due to the small samples size, such models are prone to over-fitting. 
There are three approaches: data augmentation, metric learning, and meta-learning to address this problem.
The method of data augmentation mainly involves feature extraction from the data and then augmenting the features.
Metric learning mainly learns from similarity matrix in which similar pairs of samples will have higher similarity scores.
Then the sample to be classified by finding the proximity class through calculating the distance between the sample to be classified and the known classified.
The meta-learning approach mainly hopes to train a model that can learning to learn and it can use previous knowledge for a new task learning.

The PA-CRF \cite{2020-PACRF-FSED} model used the meta-learning approach to solve few-shot event detection task. 
The problem of trigger discrepancy between event types is solved by adding conditional random fields.

The DMB-PN \cite{2020-DMBPN-FSED} model also used the meta-learning with Dynamic-Memory-Based Prototypical Network. 
Dynamic-Memory-Based Prototypical Network makes the model more robust and make it extract contextual information better.

Zheng propose TaLeM \cite{2021-TaLeM}, which is a novel taxonomy-aware learning model.
TaLeM consisting of two components, the taxonomy-aware self-supervised learning framework (TaSeLF) and the taxonomy-aware prototypical networks (TaPN). 
They use TaSeLF to mines the taxonomy-aware distance relations to increases the training examples.
It can help alleviates the generalization bottleneck brought by the insufficient data. 
They also use TaPN to introduces the Poincaré embedding to represent the label taxonomy, and integrates them into a task-adaptive projection networks.
It can help tackles problems of the class centroids distribution and the taxonomy-aware embedding distribution ins the vanilla prototypical networks.

Li propose P4E \cite{2022-P4E}, which is an identify and localize event detection framework. 
It integrates the best of few-shot prompting and structured prediction. 
This framework decomposes event detection into an identification task and a localization task. 
P4E use cloze-based prompting to align the objective with the pre-training task of language models and employ an event type-agnostic sequence labeling model to localize the event trigger conditioned on the identification output.

\subsection{Few-shot Incremental learning}
Few-shot incremental learning has mainly been studied in image recognition, where models are trained on the base class first and then generalized to new classes. 
However, the focus of few-shot incremental learning is on incremental learning, which means the model must be able to learn from few numbers of new samples without forgetting the knowledge it has already learned.

Two common approaches in few-shot incremental learning are meta-learning-based and classifier-based methods. 
The former makes the model learn how to detect event types, while the latter performs event type detection only for the learned types. 
After training with meta-learning methods, the model is capable of learning new classes in addition to predicting both new and previous classes. 

The Attention Attractor Network \cite{2018-AAN-IFSL} regularizes the learning of novel classes and facilitates the learning of these parameters.
In each episode, it trains a set of new weights to recognize novel classes until they converge.
The technique can back-propagate through the optimization process and it can also facilitate the learning of these parameters.

Sung proposes XtarNet \cite{2020-XtarNet}, which learns to extract task-adaptive representations (TAR) for facilitating incremental few-shot learning. 
This method utilizes a pre-trained backbone network on a set of base categories and additional modules that are meta-trained across episodes. 
When given a new task, the novel feature extracted from the meta-trained modules is mixed with the base feature obtained from the pre-trained model. 
This process of combining two different features provides TAR and is also controlled by meta-trained modules.

Tao redefines few-shot incremental learning and proposes the TOPIC \cite{2020-TOPIC-FSCIL} model to solve this problem. 
This paper argues that true few-shot incremental learning can only train on the increments, not on the entire data. 
The difference between this approach and the meta-learning-based approach is that it first trains on a large-scale dataset and then keeps adding new datasets with different categories from the base dataset.
When learning new samples, the base dataset is not retained, and only the new samples are trained. 
This method is closer to the concept of incremental learning than the first method, and the original model can recognize both old and new data types by training only on the new dataset, which is also more advantageous in practical applications.

Dong proposes the exemplar relation distillation incremental learning framework \cite{2021-RKD} to balance the tasks of preserving old knowledge and adapting to new knowledge. 
An exemplar relation loss function for discovering the relation knowledge between different classes is introduced to learn and transfer the structural information in the relation graph.

Xia first proposes using few-shot incremental learning on text classification in his paper \cite{2021-NRNC-IFSTC}. 
This paper defines a new task in the NLP domain, incremental few-shot text classification, where the system incrementally handles multiple rounds of new classes. 
They proposed ENTAILMENT and HYBRID to solve this novel problem.

To date, few-shot incremental learning has not been extensively explored in the context of event detection. 
Thus, we endeavor to apply this approach to the field of event detection, which in turn yields a new task, namely few-shot incremental event detection. 
We use hybrid distillation as a loss function and use sample retention to solve the problem of catastrophic forgetting in incremental learning.
External knowledge is also introduced to solve the problem of learning efficiency degradation caused by insufficient sample.
\section{Conclusion}
\label{sec:conclusion}
In this work, we proposed and defined few-shot incremental event detection task for the first time and reconstructed a dataset for it. We propose two models for this particular challenge: IFSED-K and IFSED-KP. The analysis of the results obtained on the new dataset shows that the present model achieves substantial improvement compared with the previous work. Based on the analysis of the experimental results, in the future, we consider choosing a loss function more suitable for few-shot learning to supplement the existing.




\section{Acknowledgments}
This work is supported by R\&D Program of Beijing Municipal Education Commission (KM202210009002), the National Natural Science Foundation of China (61972003) and the Beijing Urban Governance Research Center Fund. And we would also like to thank the anonymous reviewers for their helpful comments. We would like to thank the referees for their comments, which helped improve this paper considerably.

\bibliographystyle{unsrt}
\bibliography{refs}
\end{CJK*}
\end{document}